\documentclass[11pt]{article}
\usepackage{ACL2023}
\usepackage{times}
\usepackage{latexsym}
\usepackage[T1]{fontenc}
\usepackage[utf8]{inputenc}
\usepackage{microtype}
\usepackage{inconsolata}
\usepackage{comment} 
\usepackage{graphicx}
\usepackage{xcolor} 
\definecolor{lightergrey}{gray}{0.9} 
\usepackage{hyperref}
\usepackage{listings}
\usepackage{xcolor} 
\definecolor{codegreen}{rgb}{0,0.6,0}
\definecolor{codegray}{rgb}{0.5,0.5,0.5}
\definecolor{codepurple}{rgb}{0.58,0,0.82}
\definecolor{backcolour}{rgb}{0.95,0.95,0.92}
\lstdefinestyle{mystyle}{
    backgroundcolor=\color{backcolour},   
    keywordstyle=\color{codepurple},
    numberstyle=\tiny\color{codegray},
    stringstyle=\color{codegreen},
    basicstyle=\ttfamily\footnotesize,
    breakatwhitespace=false,         
    breaklines=true,                 
    captionpos=b,                    
    keepspaces=true,                 
    showspaces=false,                
    showstringspaces=false,
    showtabs=false,                  
    tabsize=2,
    framexleftmargin=8pt,
    framexrightmargin=8pt,
}
\usepackage{hyperref}
\usepackage{enumitem}
\usepackage{float} 
\usepackage{caption} 
\usepackage{{csvsimple}}

\title{Building Efficient Universal Classifiers with Natural Language Inference}

\author{Moritz Laurer\textsuperscript{\textdaggerdbl}\textsuperscript{\textbardbl}, Wouter van Atteveldt\textsuperscript{\textdaggerdbl}, Andreu Casas\textsuperscript{\textdagger}, Kasper Welbers\textsuperscript{\textdaggerdbl}\\
  \textsuperscript{\textdaggerdbl} Vrije Universiteit Amsterdam \\
  \textsuperscript{\textdagger} University of London, Royal Holloway \\
  \textsuperscript{\textbardbl} Hugging Face \\
  \texttt{moritz@huggingface.co}
}

\begin{document}

\maketitle

\begin{abstract}
Generative Large Language Models (LLMs) have become the mainstream choice for fewshot and zeroshot learning thanks to the universality of text generation. Many users, however, do not need the broad capabilities of generative LLMs when they only want to automate a classification task. Smaller BERT-like models can also learn universal tasks, which allow them to do any text classification task without requiring fine-tuning (zeroshot classification) or to learn new tasks with only a few examples (fewshot), while being significantly more efficient than generative LLMs. This paper (1) explains how Natural Language Inference (NLI) can be used as a universal classification task that follows similar principles as instruction fine-tuning of generative LLMs, (2) provides a step-by-step guide with reusable Jupyter notebooks for building a universal classifier,\footnote{\url{https://github.com/MoritzLaurer/zeroshot-classifier}} and (3) shares the resulting universal classifier that is trained on 33 datasets with 389 diverse classes. Parts of the code we share has been used to train our older zeroshot classifiers that have been downloaded more than 55 million times via the \includegraphics[width=1em]{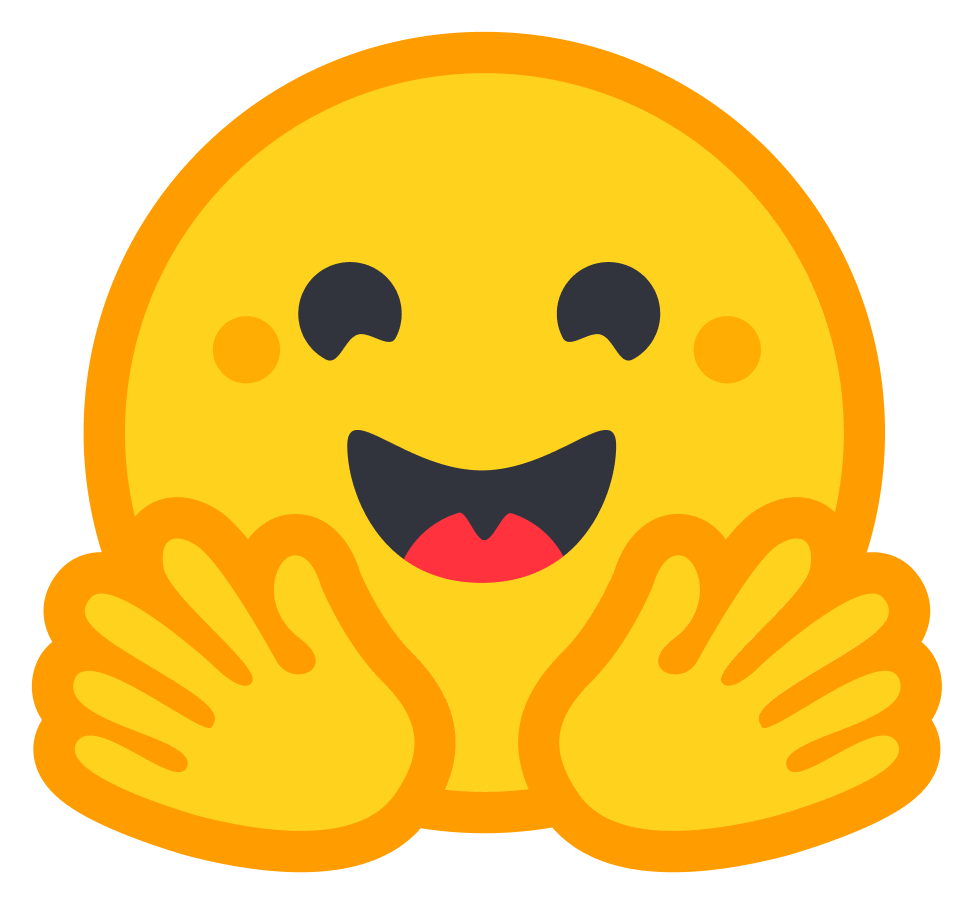} Hugging Face Hub as of December 2023. Our new classifier improves zeroshot performance by 9.4\%.
\end{abstract}

\begin{figure*}[ht]
    \centering
    \includegraphics[width=\linewidth]{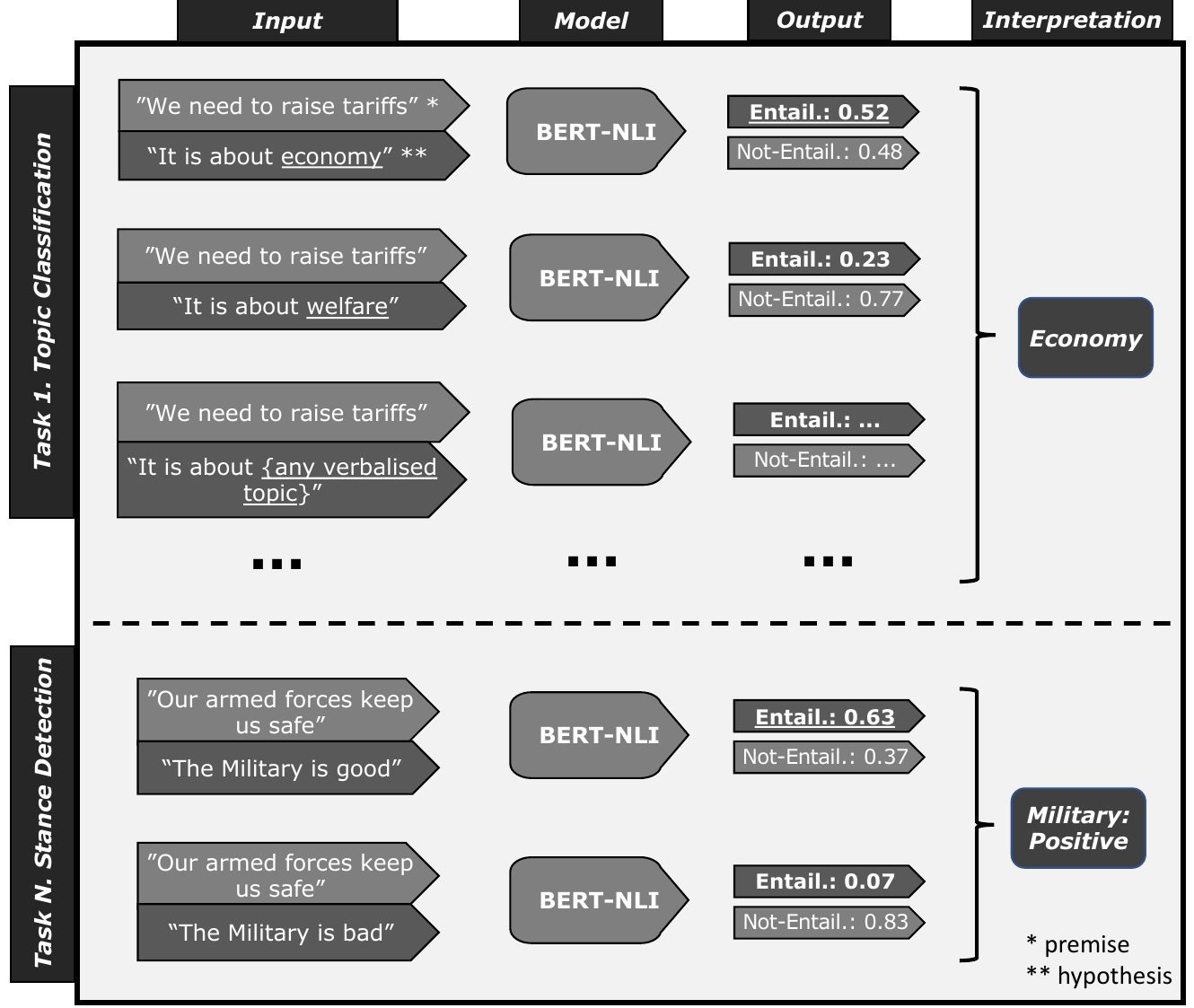}
    \caption{Illustration of universal classification with BERT-NLI based on \citealp{laurer_less_2023}}
    \label{fig:figure_bert_nli}
\end{figure*}

\section{Introduction}

Over the past year, generative models have taken both academia and public attention by storm. The main appeal of text generation is that it is so universal, that almost any other text-related task can be reformulated as a text generation task \citep{radford_language_2019,raffel_exploring_2020}. Especially when text generators are massively scaled up and tuned on human instructions, they acquire impressive capabilities to generalise to new tasks without requiring task-specific fine-tuning \citep{sanh_multitask_2022,ouyang_training_2022,chung_scaling_2022,openai_gpt-4_2023,touvron_llama_2023}. Since the utility of these generative Large Language Models (LLMs) has become evident, large amounts of intellectual, financial and energy resources are being invested in improving and scaling generative LLMs.

Given that the resource requirements for training and deploying generative LLMs are prohibitive for many researchers and practitioners, this paper investigates other types of universal models, that make a different trade-off between resource requirements and universality. The literature has developed several other universal tasks that cannot solve generative tasks (summarization, translation etc.), but can solve \textit{any} classification task with smaller size and performance competitive with generative LLMs \cite{xu_universal_2023,schick_its_2021}.


The principle of universal classifiers is similar to generative models: A model is trained on a universal task, and a form of instruction or prompt enable it to generalize to unseen classification tasks. While several efficient approaches to universal classification exist \citep{schick_exploiting_2021,xia_prompting_2022,yao_prompt_2022,xu_universal_2023,bragg_flex_2021,ma_issues_2021,sun_nsp-bert_2022}, this paper focuses on guidance for one approach: Natural Language Inference. Several papers have used the universal NLI task for zero- and fewshot classification, but stopped short of mixing NLI data with multiple other non-NLI datasets to build more universal classifiers \citep{yin_benchmarking_2019,yin_universal_2020,wang_entailment_2021,laurer_less_2023}. 

The main contribution of this paper are: (1) easy-to-use universal classifiers trained on 5 NLI datasets and 28 non-NLI datasets with 389 diverse classes, improving zeroshot performance by 9.4\% compared to NLI-only models; (2) a step-by-step guide with Juypter notebooks enabling users to train and adapt their own universal classifiers.  

\vspace{30pt}

\section{NLI as a universal task}

The Natural Language Inference (NLI) task\footnote{An older but more expressive name for the task is RTE, Recognising Textual Entailment \citep{quinonero-candela_pascal_2006}} is defined as recognising if the meaning of one text (the hypothesis) is entailed in another text (the premise). For example, the hypothesis ``The EU is not trustworthy" is entailed in the premise ``The EU has betrayed its partners during the negotiations on Sunday". To create NLI datasets, workers are presented with a text (the premise) and are tasked with writing a hypothesis that is either clearly true given the premise (entailment), clearly false given the premise (contradiction), or that might be true or false but is not clearly entailed or a contradiction (neutral). Several large scale NLI datasets with hundreds of thousands of unique hypothesis-premise pairs for these three classes have been created by crowd workers or language models \citep{bowman_large_2015,williams_broad-coverage_2018,conneau_xnli_2018,nie_adversarial_2020,parrish_does_2021,liu_wanli_2022}. For simplicity and to increase universality, the task can be simplified into a binary entailment vs. not-entailment task by merging the `neutral' and `contradiction' labels \citep{yin-etal-2021-docnli}.

\begin{figure*}[!t]
    \begin{lstlisting}[language=Python]
from transformers import pipeline

text = "Angela Merkel is a politician in Germany and leader of the CDU"
hypothesis_template = "This text is about {}"
classes_verbalized = ["politics", "economy", "entertainment", "environment"]

model_name = "MoritzLaurer/deberta-v3-large-zeroshot-v1.1-all-33"
classifier = pipeline("zero-shot-classification", model=model_name)
classifier(text, classes_verbalized, hypothesis_template=hypothesis_template)

# output: {'labels': ['politics', 'entertainment', 'economy', 'environment'], 'scores': [0.99, 0.0, 0.0, 0.0]}
\end{lstlisting}
    \caption{Example for using the resulting universal classifiers in the \includegraphics[width=1em]{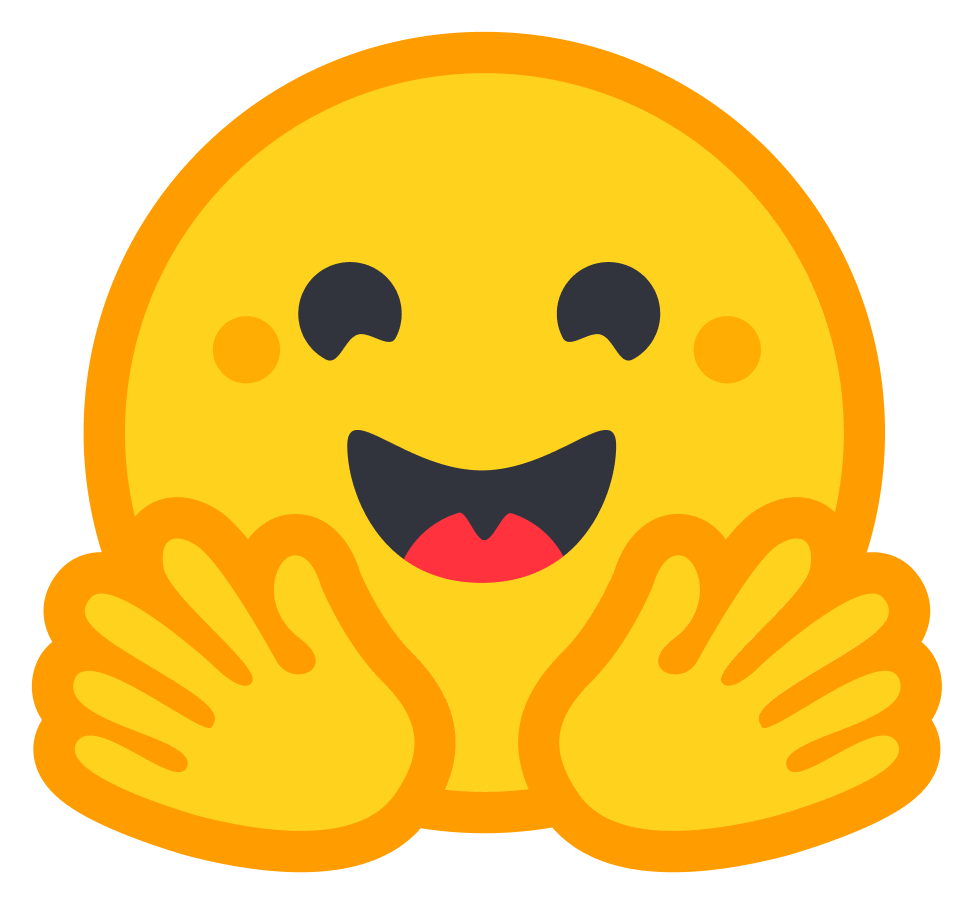} zeroshot pipeline}
    \label{fig:figure_zeroshot_pipeline}
\end{figure*}

This binary NLI task is universal, because any text classification task can be reformulated into this entailment vs. not-entailment decision through label verbalisation (see figure \ref{fig:figure_bert_nli}). Take topic classification as an example. The task could be to determine if the text ``We need to raise tariffs" belongs to the topic ``economy" or ``welfare". From an NLI perspective, we can interpret the text ``We need to raise tariffs" as the premise and verbalise the topic labels in two topic hypotheses: ``This text is about economy" and ``This text is about welfare". The classification task reformulated as an NLI task then consists of determining which of the two topic hypotheses is more entailed in the text of interest (premise). In different words: Which hypothesis is more consistent with the text of interest?

A model fine-tuned on NLI data (e.g. ``BERT-NLI") can then be used to test any hypothesis formulated by a human against any text of interest (premise). For each individual hypothesis-premise pair, an NLI models will output a probability for entailment and not-entailment. To choose the most probable topic, we can select the hypothesis with the highest entailment score. Following the same procedure, any other text classification task can be reformulated as an NLI task, from stance detection, sentiment classification to factuality classification (see figure \ref{fig:figure_bert_nli}). Any class can be verbalised as a hypothesis (similar to the prompt of a generative LLM) and can then be tested against any text of interest.\footnote{Note that an NLI model will always only do one task (NLI) just like a GPT model can only predict the next token. These tasks are universal because any other specific task can be reformatted into these more general tasks.} 

The main disadvantage of NLI for universal classification is that it requires a separate prediction for each of N class hypotheses, creating computational overhead for tasks with many classes. The main advantage is that identifying a new class only requires verbalising it as a hypothesis and passing it to an NLI model without the need of fine-tuning a new task-specific model from scratch (zeroshot classification). The most prominent implementation of this approach is probably the Hugging Face \texttt{ZeroShotClassificationPipeline} (see figure \ref{fig:figure_zeroshot_pipeline}) which uses this NLI-based approach under the hood \citep{wolf_transformers_2020}.\footnote{\url{https://huggingface.co/docs/transformers/v4.21.2/en/main_classes/pipelines##transformers.ZeroShotClassificationPipeline}} The models created in the paper are designed to be directly compatible with this pipeline.

\section{A guide to building a universal classifier}

In this guide we explain how this type of universal classifier is built. Each step is accompanied by a Jupyter notebook available on GitHub that implements each step end-to-end.\footnote{\url{https://github.com/MoritzLaurer/zeroshot-classifier}} The main steps are: 
\begin{enumerate}[itemsep=0pt, parsep=0pt]
  \item Dataset preprocessing and harmonization
  \item Automatic data cleaning (optional)
  \item Hypothesis formulation and formatting
  \item Training and evaluation
  \item Visualisation of results
\end{enumerate}

Guidance for using the resulting model is provided in section 4.

\subsection{Data selection, preprocessing and harmonization}
We use two main types of data to train our universal classifier: Five NLI datasets and 28 other classification datasets. 

\subsubsection{\centering\texttt{\href{https://github.com/MoritzLaurer/zeroshot-classifier/blob/main/v1_human_data/1_data_harmonization_nli.ipynb}{data-harmonization-nli.ipynb}}}

First, we use a set of established NLI datasets: MNLI \citep{williams_broad-coverage_2018}, ANLI, FEVER-NLI \citep{nie_adversarial_2020}, WANLI \citep{liu_wanli_2022}, Ling-NLI \citep{parrish_does_2021}.\footnote{We exclude the large SNLI datasets \cite{bowman_large_2015} due to known issues of data quality.} Each dataset contains tens of thousands of unique hypothesis-premise pairs classified into one of the three classes ``entailment", ``neutral", ``contradiction".  We merge the ``neutral" and ``contradiction" class into one ``not-entailment" class to obtain the universal binary format. As figure \ref{fig:figure_bert_nli} shows, only the probabilities for the ``entailment" class are relevant for universal classification. We merge all five NLI datasets into one harmonized dataset with three columns: ``premise", ``hypothesis", ``label". 

The resulting merged \~{}885000 hypothesis-premise pairs would be enough to train a decent NLI model capable of zeroshot classification. The NLI datasets were, however, not created with zeroshot classification in mind. Crowd workers were instructed to write hypotheses that are entailed, contradictory or neutral towards a text, which led to a wide range of hypothesis-premise pairs. They were not specifically instructed to create data for typical classification tasks such as identifying topics, sentiment, stances, emotions, toxicity, factuality etc. which users might be interested in in practice (e.g. ``This text is about topic X"). To improve performance on these types of tasks, we therefore add a second collection of standard non-NLI classification datasets reformatted into the NLI format.

\subsubsection{\centering\texttt{\href{https://github.com/MoritzLaurer/zeroshot-classifier/blob/main/v1_human_data/1_data_harmonization_huggingface.ipynb}{data-harmonization-huggingface.ipynb}}}

We choose 28 popular non-NLI datasets with diverse classification tasks linked to sentiment, emotions, intent, toxicity, bias, topics, factuality, spam etc. with 387 classes in total. We selected most datasets based on their popularity (downloads) on the Hugging Face Hub. We also add some non-NLI datasets that are not available on the Hugging Face hub and create separate preprocessing notebooks for each of them (e.g. \texttt{1-data-harmonization-manifesto.ipynb}). The full list of datasets with information on tasks, licenses and data quality is available in our dataset overview file.\footnote{\url{https://github.com/MoritzLaurer/zeroshot-classifier/blob/main/v1_human_data/datasets_overview.csv}} 

For creating this kind of collection, we strongly recommend manually inspecting each dataset and the corresponding paper to understand data quality and the underlying task. Depending on the datasets, the preprocessing steps can include: removing NAs, deduplication, downsampling majority classes, merging texts (e.g. titles with text bodies), converting continuous labels into simpler classes (e.g. star ratings to binary sentiment classes), removing texts with low certainty or annotator agreement, splitting datasets with multiple implicit tasks into separate tasks, removing and renaming columns, and splitting the data into a 80-20 train-test split if no test-set exists. As a result of these steps, each processed dataset only has three harmonized column: ``text", ``label\_text" (a word expressing the meaning of each class), and ``label\_standard" (a number for each class). 

If readers want to improve the classifier on a specific domain or a family of other tasks, they can add their datasets during this step.

\subsection{Automatic data cleaning}

\subsubsection{\centering\texttt{\href{https://github.com/MoritzLaurer/zeroshot-classifier/blob/main/v1_human_data/2_data_cleaning.ipynb}{data-cleaning.ipynb}}}

Manual inspection of the non-NLI datasets reveal relevant quality issues in many datasets. We therefore use the \texttt{CleanLab} library to remove texts with a high probability of noise.\footnote{\url{https://github.com/cleanlab/cleanlab}} \texttt{CleanLab} provides automated means for identifying noisy labels by embedding texts with a SentenceBERT model, training a simple logistic classifier on these embeddings and analysing prediction uncertainty and prediction overlaps between classes. 

Two relevant limitations of this process are that it can disproportionately remove minority classes and it probably does not work well for very complex tasks. We therefore applied this automatic approach to 25 tasks, but not to complex tasks like NLI or factuality detection. This process removes roughly 17\% (or \~{}135 000) texts with probable misclassifications or label overlaps. We highly recommend readers to inspect our cleaning notebook to get a feeling for the amount of noise that is still present in established datasets. 

As an additional measure to increase data quality and diversity in the following script, we also radically downsample data for each non-NLI dataset. We only take a sample of maximum 500 texts per class and maximum 5000 texts per dataset to avoid overfitting to a specific large dataset. This leads to 51731 non-NLI texts (down from more than one million texts) that will be merged with the \~{}885000 NLI texts in the following step. We could have added hundreds of thousands of additional texts, but our experience indicates that data diversity and quality is more important than quantity. Moreover, our objective is not to build a classifier that beats (and overfits to) a benchmark, but to build a classifier that generalizes well.

\subsection{Hypothesis formulation and NLI formatting}

\subsubsection{\centering\texttt{\href{https://github.com/MoritzLaurer/zeroshot-classifier/blob/main/v1_human_data/3_data_formatting_universal_nli.ipynb}{data-formatting-universal-nli.ipynb}}}

We now need to transform the (cleaned) non-NLI datasets into the universal NLI format. First, we need to verbalise each class as a class hypothesis. For this label verbalisation step we read the underlying paper or annotator instructions for each dataset and express them as a class hypothesis. For a binary sentiment classification task on app reviews, for example, the hypotheses could be ``This app review text expresses positive sentiment" and ``This app review text expresses negative sentiment". We add information on the domain or type of dataset (``app review text") in some hypotheses, to help the model differentiate between texts from the same task type (e.g. binary sentiment classification) that come from different domains or datasets (e.g. app reviews vs. movie reviews vs. product reviews). This helps reduce negative transfer risks across datasets. As a general rule, we try to formulate the hypotheses in simple every-day language and avoid complex academic definitions, thinking of the model a bit like a simple crowd worker. Each class hypothesis is linked to its corresponding class label in a dictionary. All our hypotheses are available in \texttt{3-data-formatting-universal-nli.ipynb}.\footnote{Research indicates that providing multiple different instructions (hypotheses) for the same class can help increase generalisation \citep{sanh_multitask_2022}.}

For each row in each non-NLI \textit{training} dataset we now add a new ``hypothesis" column with the correct class hypothesis corresponding to the respective text. Moreover, in a new ``label" column, these text-hypothesis pairs receive the label ``0" for ``entailment". We then multiply each text by two and pair the copied text with a random incorrect class hypothesis and the label ``1" for ``not-entailment". This multiplication ensures that the model does not only learn that class hypotheses are always true and it functions as a form of data augmentation. When we rename the ``text" column to ``premise", this dataset now has exactly the same format as the NLI dataset with the columns ``premise", ``hypothesis", ``label" for binary entailment vs. not-entailment classification. This conversion is implemented in the function \texttt{format\_nli\_trainset}. We can now simply concatenate the non-NLI and the NLI training data.

The non-NLI \textit{test} data needs to be formatted slightly differently. During test-time, all class hypotheses for a task need to be tested on each text to select the ``most entailed" hypothesis. This means that we need to multiply each test text by N for N classes, pairing the text with all N possible class hypotheses in N rows. This conversion is implemented in the function \texttt{format\_nli\_testset}. After this task-specific multiplication, these test sets cannot be concatenated and they need to be evaluated separately.

\subsection{Training and evaluation}

\subsubsection{\centering\texttt{\href{https://github.com/MoritzLaurer/zeroshot-classifier/blob/main/v1_human_data/4_train_eval.ipynb}{train-eval.ipynb}}}

With the data fully cleaned and formatted, we can now start training. We can use any pre-trained transformer model as the foundation. Since the only purpose of the model is classification, we discard models with a decoder such as T5 or Llama-2 \cite{raffel_exploring_2020,touvron_llama_2023}. Among encoder-only models, we had the best experience with DeBERTaV3 which is pre-trained with the highly effective RTD objective and exists in multiple sizes and with a multilingual variant \citep{he_debertav3_2021}. Processing and training is implemented with Hugging Face \texttt{Transformers}. We use \texttt{label2id = \{"entailment": 0, "not\_entailment": 1\}} for compatibility with the \texttt{ZeroShotClassificationPipeline}; pad and truncate to a maximum length of 512 tokens; base hyperparameters on the recommended fine-tuning hyperparameters in the appendix of the DeBERTaV3 paper \cite{he_debertav3_2021} and do not conduct a hyperparameter search as it adds little value over the recommended hyperparameters in our experience while adding complexity. 

We fine-tune models with three different data compositions for evaluation: (1) one model trained on all datasets (\texttt{deberta-v3-zeroshot-v1.1-all-33}); (2) one model trained on only the five NLI datasets as a baseline representing previous NLI-only zeroshot models (\texttt{deberta-v3-nli-only}); (3) 28 different models, each trained with all datasets, except one non-NLI dataset is held out. This last group of models is trained to test zeroshot generalisation to tasks the model has not seen during training. For each of the 28 models, we take the performance metric for the dataset that was held out in the respective training run. Based on these 28 metrics, we know what the performance for each task would be, if the model had seen all datasets, except the respective held out dataset. 

One training run on around 9000000 concatenated hypothesis-premise pairs for 3 epochs takes around 5 hours for DeBERTaV3-base and 10 hours for DeBERTaV3-large on one A100 40GB GPU. Training and evaluating all 30 models takes around 6 (base) or 15 (large) full days of compute, mostly due to the the 28 models trained for held-out testing.

We use balanced accuracy as our main evaluation metric \citep{buitinck_api_2013} as many of our datasets are class imbalanced and the metric is easier to interpret than F1 macro. For evaluation on non-NLI datasets, remember that rows have been multiplied with one row per class hypothesis. The \texttt{compute\_metrics\_nli\_binary} function handles the calculation of metrics for these reformatted datasets.

\texttt{deberta-v3-zeroshot-v1.1-all-33} is the model we recommend for downstream use. The model is available in different sizes in our zeroshot collection on the Hugging Face Hub.\footnote{\url{https://huggingface.co/collections/MoritzLaurer/zeroshot-classifiers-6548b4ff407bb19ff5c3ad6f}}

\subsection{Visualisation and interpretation of results}

\subsubsection{\centering\texttt{\href{https://github.com/MoritzLaurer/zeroshot-classifier/blob/main/v1_human_data/5_viz.ipynb}{viz.ipynb}}}

\begin{figure}[h]
    \centering
    \includegraphics[width=\linewidth]{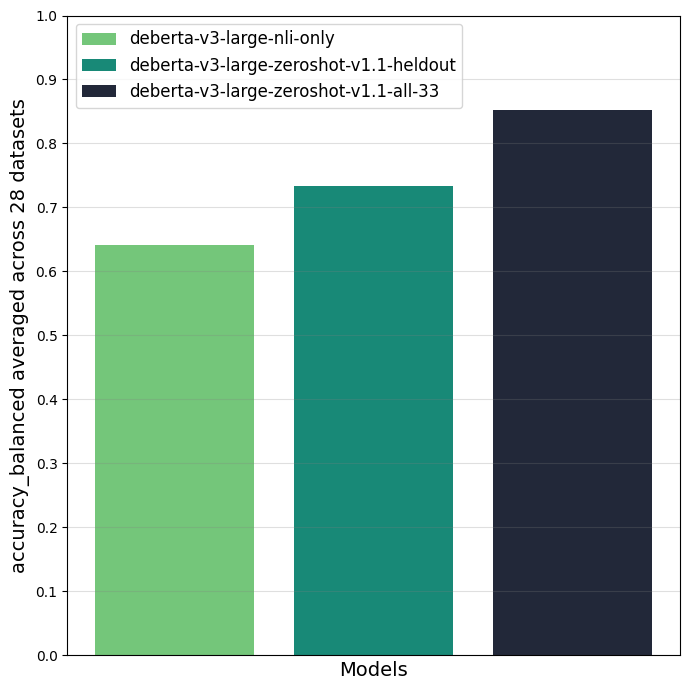}
    \caption{Mean performance across 28 classification tasks. }
    \label{fig:figure_results_large_average}
\end{figure}

The NLI-only classifier (\texttt{deberta-v3-nli-only}) is very similar to existing zeroshot classifiers on the Hugging Face hub. It can do all tasks to some extent, given it's training on the universal NLI task. It performs well on simple binary tasks such as sentiment classification, but struggles on other tasks that are too dissimilar from standard NLI texts and have more classes. 

\texttt{deberta-v3-zeroshot-v1.1-all-33} has seen up to 500 examples for each class in each dataset. Only based on this small amount of data, it achieves strongly improved performance across all tasks. This is in line with prior research indicating that little, but good quality data is necessary for language models to generalize well \citep{zhou_lima_2023}.

\texttt{deberta-v3-zeroshot-v1.1-heldout} provides an indication of zeroshot performance for tasks the model has not seen during training. We highlight two main insights: First, models trained with a mix of NLI data and non-NLI data achieve overall better zeroshot performance than the NLI-only model (+9.4\% on average). Having seen different zeroshot-style hypotheses helps the model generalize to other unseen tasks and hypotheses (positive transfer). Second, there are a few cases of negative transfer. On a few datasets, the NLI-only model performs better than \texttt{deberta-v3-zeroshot-v1.1-heldout}, indicating that the additional task-mix can make the model over- or underpredict a few classes. 

Overall, \texttt{deberta-v3-zeroshot-v1.1-all-33} significantly outperforms the NLI-only model both on held-in and held-out tasks. Its performance on datasets it has not seen during training can expected to be around 9.4\% higher than NLI-only models. Moreover, it can simultaneously perform many different tasks it has seen during training with even better performance. Detailed metrics are available in the appendix and the model cards.\footnote{\url{https://huggingface.co/MoritzLaurer/deberta-v3-large-zeroshot-v1.1-all-33}}

\section{Reusing our models and code}

We envisage three main ways in which our models and code can be reused. First, users can directly use \texttt{deberta-v3-zeroshot-v1.1-all-33} for zeroshot classification in just a few lines of code with the Hugging Face \texttt{ZeroShotClassificationPipeline} (see code in figure \ref{fig:figure_zeroshot_pipeline}). This should work particularly well for tasks that are similar to one of the 33 datasets and 389 classes we used for training, including many different topics, sentiment, emotions, or types of toxicity. 

Second, the models can be used as a base models to fine-tune a task-specific classifier. Prior research shows that fine-tuning an NLI-based classifier requires less training data and increases robustness compared standard fine-tuning of DeBERTaV3-base \citep{laurer_less_2023,raman_model-tuning_2023,le_scao_how_2021}. Good performance can be achieved with just a few hundred examples per class, requiring only some minutes of fine-tuning on a free GPU \citep{laurer_lowering_2023}. We provide code examples for this approach in an online workshop.\footnote{See the notebook \texttt{4\_tune\_bert\_nli.ipynb} at \url{https://github.com/MoritzLaurer/summer-school-transformers-2023/tree/main}}.

Third, researchers can modify our notebooks, for example by adding more datasets for a specific domain and task family, and rerun the improved pipeline to build a universal classifier that is better adapted to their domain and tasks. While fine-tuning \texttt{deberta-v3-zeroshot-v1.1-all-33} is recommended for individual tasks, rerunning the pipeline could add value if researchers want to build a new universal model adapted to a broader set of tasks or domains. We estimate that the final model can be trained with a € 50 Google Colab Pro+ subscription. 

In all three use-cases, making predictions with the resulting models (inference) is highly efficient with cheap GPUs, but is also possible with on a laptop CPU.

\section{Limitations}

We outline several limitations of this paper and invite readers to improve on our implementation. First, while we have included 28 non-NLI datasets, the diversity of these academic datasets is limited and they do not cover the full diversity of classification use-cases users will need in practice.  All datasets are only in English. The instruction fine-tuning literature for generative LLMs has shown the potential of using SotA models like GPT-4 to generate diverse training data and distilling their capabilities into much smaller models \citep{taori_alpaca_2023,tunstall_zephyr_2023}. While many such datasets exist for generative tasks, hardly any are available for encoder-only classifiers like BERT \citep{sileo_tasksource_2023,longpre_flan_2023,longpre_data_2023}. We assume that smart LLM prompting could result in a more diverse dataset than our collection and could further improve generalisation. 

Second, the model comparisons are limited as we only compare BERT-NLI models among each other. We do not compare classification performance, inference speed, memory requirements, and costs to larger generative LLMs or APIs.

Third, we assume that our data still contains a certain degree of noise. Additional data cleaning techniques could be used, for example discarding training data where the DeBERTa-v3 model still disagrees with the label after fine-tuning or targeted manual inspection enabled by active learning. 

Fourth, an inherent limitation of NLI for zeroshot classification is that each additional class requires an additional forward pass (prediction) through the model. This makes the approach less suitable for tasks with a high amount of classes. At the same time, even if multiple forward passes are required, encoder-only models with only around a hundred million parameters are still more efficient than decoder models with multiple billion parameters while possibly being more accurate \citep{xu_universal_2023,schick_its_2021}. 

Fifth, we use the relatively old DeBERTa-v3 from November 2021 \citep{he_debertav3_2021}, which misses relevant recent innovations like longer context windows or flash attention \citep{dao_flashattention_2022}. Unfortunately we are not aware of a better encoder-only model and releases have recently been dominated by larger generative decoder models. 

Sixth, several other universal classification approaches exist that were beyond the scope of this paper: PET, which combines masked-language-modeling and label verbalisation \citep{schick_exploiting_2021}, replaced-token-detection combined with prompts \citep{xia_prompting_2022,yao_prompt_2022,xu_universal_2023}, question-answering \citep{bragg_flex_2021}, or next-sentence-prediction as an interesting self-supervised alternative to NLI \citep{ma_issues_2021,sun_nsp-bert_2022}.

\section{Conclusion and call for a new foundation model}

This paper explains how to use the Natural Language Inference task to build a universal classifier and provides practical guidance to users. Looking forward, we believe that there is significant room for improvement by building upon the insights from generative LLM research for more efficient classifiers. 

First, generative LLMs gain their power by learning their universal task (next-token-prediction) already during self-supervised pre-training and not only during fine-tuning (a limitation of our models). It is possible that universal self-supervised tasks exist for classification tasks as well (or discriminative tasks more generally). The most promising candidate is ELECTRA's replaced-token-detection (RTD) objective \citep{clark_electra_2020}, which can make models with only a few hundred million parameters perform comparably to models with 1.5 billion parameters that are trained on the the less efficient generative masked-language-modeling objective \citep{he_debertav3_2021}. We hypothesize that the RTD objective could be supplemented with a binary ``original text" vs. ``not-original text" objective, resulting in a universal classification head similar to the universal ``entailment" vs. ``not-entailment" task - without requiring supervision. \citet{xu_universal_2023} go in this direction, but did not experiment with a self-supervised task.

Second, a new foundation model trained on this task could then also be trained with other more recent innovations, which existing encoder-only models are currently lacking: flash attention \citep{dao_flashattention_2022}, grouped-query attention \citep{ainslie_gqa_2023}, better positional embeddings like RoPe or AliBi to enable longer context windows \citep{su_roformer_2023,press_train_2022}, and scaling pre-training data and compute while only moderately scaling model size for inference-time efficiency \citep{hoffmann_training_2022}.

Third, similar to generative LLMs, better instruction data could make universal classifiers more useful. As discussed in the limitations section, especially synthetic data from much larger generative LLMs tailored to universal classifiers has the potential to flexibly teach efficient classifiers more diverse and more practically relevant tasks. The creators of the WANLI dataset have already demonstrated this potential with GPT3 \citep{liu_wanli_2022} and it is safe to assume that newer generators will produce even better data. 

These points would entail pre-training a new foundation model from scratch, which requires large amounts of resources. We believe that such a foundation model for text classification would be a useful addition to the open-source ecosystem as the field has progress significantly since the last encoder-only models were released and classification tasks constitute a relevant share of both academic and practical applications for language models.

\vspace{30pt}

\section*{Acknowledgements and funding}
This work was supported by a PhD scholarship of the Heinrich Böll Foundation and by a Snellius compute grant (EINF-3006) by the Dutch Research Council (NWO). The authors would like the thank Camille Borrett for her feedback during the drafting process.

\vspace{40pt}

\bibliography{anthology,custom}
\bibliographystyle{acl_natbib}

\newpage
\appendix

\section{Metrics details}
\label{sec:appendix_a}

Detailed metrics per dataset are reported in the model cards for the base-sized model at \url{https://huggingface.co/MoritzLaurer/deberta-v3-base-zeroshot-v1.1-all-33} and for the large-sized model at \url{https://huggingface.co/MoritzLaurer/deberta-v3-large-zeroshot-v1.1-all-33}. See also figures \ref{fig:figure_results_large} and \ref{fig:figure_results_base} below.

\begin{figure*}
    \centering
    \includegraphics[width=\linewidth]{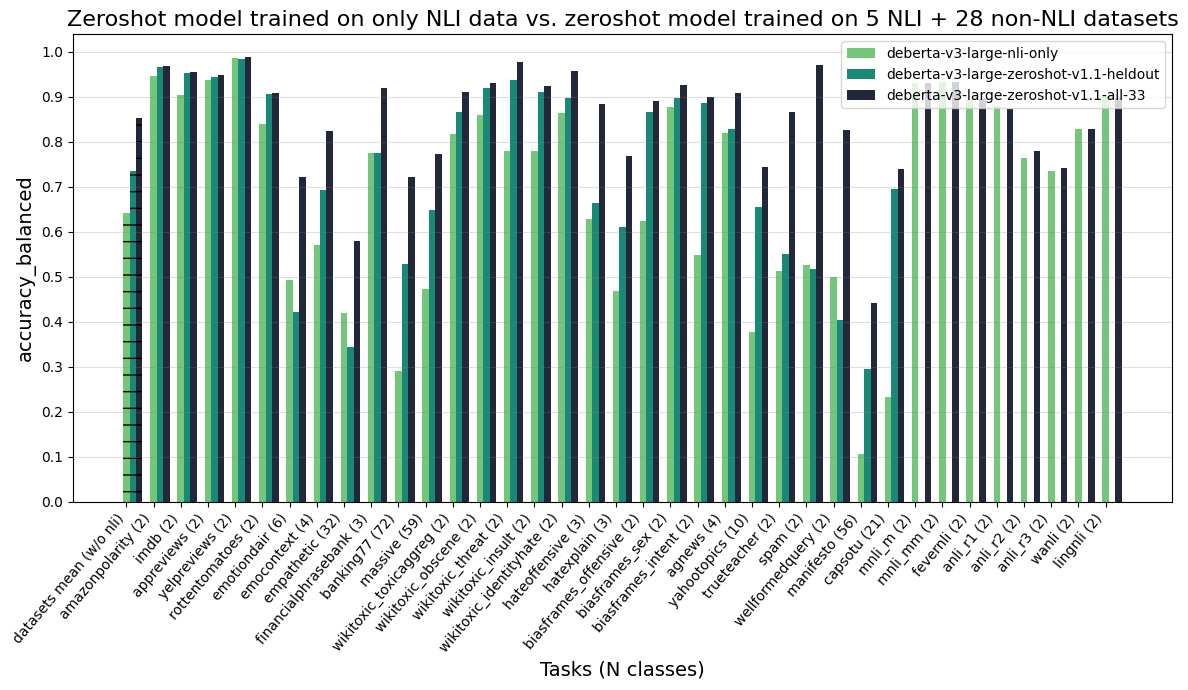}
    \caption{Metrics for large-sized model}
    \label{fig:figure_results_large}
\end{figure*}

The \texttt{financialphrasebank} dataset provides a good example for negative transfer that is present for the base sized model, but not the large model. \texttt{financialphrasebank} is a three class sentiment classification task with a third neutral category. The task mix includes other binary sentiment tasks with only the classes ``positive" vs ``negative". We assume that the base-sized model underpredicted the ``neutral" class on \texttt{financialphrasebank} under the heldout condition, as it was not sufficiently represented in the remaining data. This presumable led to a negative transfer where the NLI-only model performed better without the additional task-mix. 

\begin{figure*}
    \centering
    \includegraphics[width=\linewidth]{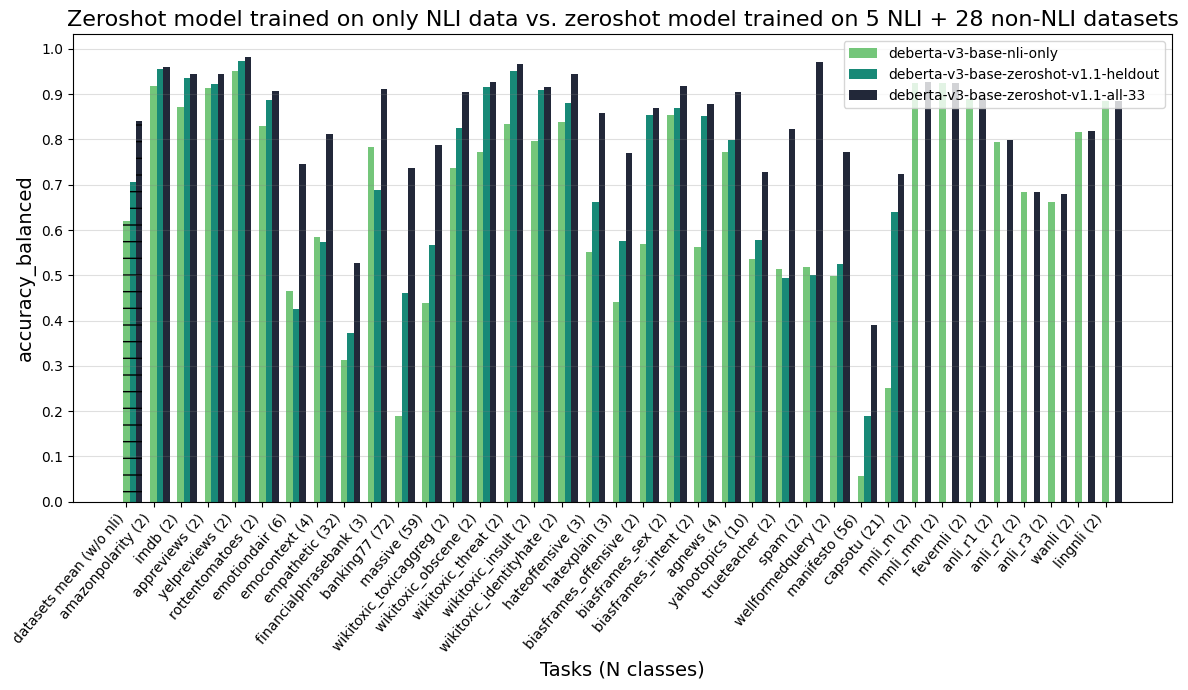}
    \caption{Metrics for base-sized model}
    \label{fig:figure_results_base}
\end{figure*}

\section{Hypotheses per task}

The exact hypotheses used for each task and class is available in the notebook \texttt{\href{https://github.com/MoritzLaurer/zeroshot-classifier/blob/main/v1_human_data/1_data_harmonization_nli.ipynb}{data-harmonization-nli.ipynb}} or in the model cards on the Hugging Face Hub:
\url{https://huggingface.co/MoritzLaurer/deberta-v3-large-zeroshot-v1.1-all-33}. For optimal performance, we recommend that users formulate their hypotheses in a similar fashion.

\section{Datasets}
For details on all datasets used, see the overview table.\footnote{\url{https://github.com/MoritzLaurer/zeroshot-classifier/blob/main/v1_human_data/datasets_overview.csv}} To give citation credit to the authors of all datasets, here is the full list of dataset sources: \citet{grano_android_2017,davidson_automated_2017,saravia_carer_2018,zhang_character-level_2015,almeida_contributions_2011,casanueva_efficient_2020,malo_good_2014,mathew_hatexplain_2021,mcauley_hidden_2013,soups_yelp_2015,faruqui_identifying_2018,maas_learning_2011,fitzgerald_massive_2023,pang_seeing_2005,chatterjee_semeval-2019_2019,sap_social_2020,rashkin_towards_2019,adams_toxic_2017,gekhman_trueteacher_2023,unknown_yahoo_answers_topics_2024,parrish_does_2021,nie_adversarial_2020,williams_broad-coverage_2018,liu_wanli_2022,burst_manifesto_2020,policy_agendas_project_us_2015,thorne_fever_2018}

\end{document}